\pdfoutput=1
\documentclass[letterpaper]{article}
\usepackage{aaai}
\usepackage{times}
\usepackage{helvet}
\usepackage{courier}
\usepackage{color}
\usepackage{comment}
\usepackage{microtype}
\usepackage{amsfonts}
\usepackage{amsmath}
\usepackage{graphicx}
\usepackage{algorithm}
\usepackage{algpseudocode}
\usepackage{subcaption}
\definecolor{darkpastelgreen}{rgb}{0.01, 0.75, 0.24}

\newcommand{\mb}[1]{\mathbf{#1}}

\algrenewcommand\algorithmicindent{1em}
\DeclareMathOperator*{\argmin}{\arg\!\min}

\title{Active Perceptual Similarity Modeling with Auxiliary Information}
\author{Eric Heim\footnotemark[1]\ \ \ \ \ \ Matthew Berger\footnotemark[2]\ \ \ \ \ \ Lee M. Seversky\footnotemark[2]\ \ \ \ \ \ Milos Hauskrecht\footnotemark[1] \\ \\
\footnotemark[1] University of Pittsburgh, Department of Computer Science, Pittsburgh, PA \\
\footnotemark[2] Air Force Research Laboratory, Information Directorate, Rome, NY
}

\begin{document}
\maketitle

\begin{abstract}
\begin{quote}
Learning a model of perceptual similarity from a collection of objects is a fundamental task in machine learning
underlying numerous applications.
A common way to learn such a model is from relative comparisons in the form of triplets: responses to
queries of the form ``Is object \emph{a} more similar to \emph{b} than it is to \emph{c}?''.
If no consideration is made in the determination of which queries to ask, existing similarity learning methods can
require a prohibitively large number of responses.
In this work, we consider the problem of \emph{actively} learning from triplets -- finding which
queries are most useful for learning.
Different from previous active triplet learning approaches, we incorporate \emph{auxiliary information}
into our similarity model and introduce an active learning scheme to find queries that are informative
for quickly learning both the relevant aspects of auxiliary data and the directly-learned similarity components.
Compared to prior approaches, we show that we can learn just as effectively with much fewer queries.
For evaluation, we introduce a new dataset of \emph{exhaustive} triplet comparisons obtained from humans and demonstrate
improved performance for different types of auxiliary information.
\end{quote}
\end{abstract}

\section{Introduction}
\label{sec:Intro}

Modeling similarity amongst a collection of objects is fundamental to many machine learning tasks.
Similarity is essential for such problems as classification, clustering, and retrieval, as it
provides a means to compare, separate, and relate objects.
Recent works have demonstrated the benefit of learning \emph{perceptual similarity}, where the goal is
to learn similarity based on human judgments.
Perceptual similarity forms an important component to such methods as fine-grained visual categorization~\cite{wah2014similarity}
and perceptual visualization design~\cite{demiralp2014learning}.

For learning perceptual similarity, previous work~\cite{kendall1990rank,stewart2005absolute} has shown the benefit of eliciting similarity
information in the form of \emph{relative comparisons}, namely triplet responses gained from answers to queries of the form: ``Is object \emph{a} more similar to \emph{b} than it is to \emph{c}?''.
Triplet queries are intuitive for humans to answer and informative for learning algorithms, as evidenced by the
recent research in relative comparison similarity learning~\cite{agarwal2007generalized,van2012stochastic}.
The effectiveness of these methods is highly dependent on the choice of queries for which responses are provided.  
For instance, if randomly selected triplet responses are supplied,
then it is likely to require almost all $O(n^3)$ responses for these methods to learn a complete model of similarity~\cite{jamieson2011low}. 
To obtain such a large amount of feedback from humans is impractical for many applications.
Hence, in an effort to reduce the number of queries to pose, recent methods have considered \emph{actively}
selecting triplet queries~\cite{tamuz2011adaptively,jamieson2011low} that have the potential to elicit the most informative feedback about how objects relate.

These active selection schemes tend to limit themselves to learning perceptual similarity without any prior knowledge about how objects relate. 
Yet, in many scenarios, it is common to have auxiliary information associated with the objects 
that captures some latent aspect of the perceptual similarity space.
In images for instance, humans must consider objects visually, and thus visual features can play a role in how humans make comparisons.
Auxiliary visual features, such as those learned from convolutional neural networks on the images themselves~\cite{donahue2014decaf}, are readily available and can be used as a basis in learning similarity.
If these features reflect how humans perceive the objects, then a learned similarity model based on features can capture a more complete notion of similarity from fewer responses.

Knowing a priori which features model perceptual similarity well is difficult in many task domains.
Thus, a significant challenge lies in identifying which portions of the feature space to use for similarity, 
as it is unlikely for the feature space to exactly conform to perceptual similarity.
In the context of active learning, we would like to pose queries which quickly help us find relevant features
from auxiliary information, as well as directly learn portions of the similarity space which are not well-represented
by features.

In this work, we introduce a novel probabilistic model that utilizes auxiliary information to learn a representation of objects which agrees with triplet responses, and to reason about the usefulness of potential queries so that active selection can be performed.
In our method, similarity is modeled as an embedding of the objects, where one portion of the embedding
is a parametric combination of auxiliary features, while the other portion is learned directly from triplet responses to resolve remaining similarity details.
Our active learning method finds queries which mutually benefit both the parametric component -- quickly finding the relevant feature
space whose generalization matches human judgment -- as well as the nonparametric component.
In evaluating our method, we introduce a dataset containing an \emph{exhaustive} set of triplet responses via crowdsourcing in order to provide
the entire pool of potential query responses for evaluation of active selection schemes.
Experimentally we demonstrate that our method can effectively learn similarity with comparable accuracy to existing schemes, using only a fraction of responses.

\section{Related Work}
\label{sec:RelWork}

Recent techniques in learning similarity from relative triplet feedback can be divided into two categories: nonparametric and parametric.
Nonparametric methods~\cite{agarwal2007generalized,tamuz2011adaptively,van2012stochastic} attempt to learn a model using only triplet responses as input.  
Typically, these methods learn an embedding of objects, or equivalently, a kernel matrix directly modeling object correlation. 
Parametric methods~\cite{mcfee2011learning,huang2011generalized} learn a distance metric parameterized by a positive semidefinite matrix, using both triplet responses and a predefined representation of the objects.
In~\cite{heim2014relative}, the authors combine these two methodologies to create a framework to learn kernel matrices.
However, their framework requires solving a semidefinite program, which is prohibitively expensive in an active learning setting, where
it is necessary to learn a kernel iteratively each time new responses are received.

In active learning~\cite{settles2010active},
 the goal is to select the most informative instances in which to learn from in order to learn an accurate model from as little supervision as possible.
Two methods have been developed for active triplet query selection.  
First, \cite{jamieson2011low} introduced a method which casts constraints from triplet responses as intersections of half spaces in a $d$-dimensional space.
Given a valid embedding, the method determines if a query is ambiguous if both possible responses result in half spaces whose intersections
with the other constraints lead to non-empty cells -- i.e. it cannot infer the triplet from the current constraints.
In contrast, Adaptive Crowd Kernel Learning (A-CKL)~\cite{tamuz2011adaptively} selects queries by how likely they are to reduce uncertainty in the positions of each object in the learned embedding.
In this work, we extend A-CKL to the case where objects have auxiliary information regarding their relationships and show that by doing so we can learn a more complete model of similarity from fewer triplet responses.

  
\section{Methodology}
\label{sec:Methodology}

We begin this section by formalizing the problem of learning similarity from relative triplet feedback, highlighting prior work that provides a basis for our technique.  
Then, we introduce our similarity learning method that uses both triplet responses and auxiliary features.  
Finally, we discuss our strategy to actively select triplet queries using our learned model.

\subsection{Learning Similarity from Triplet Responses}
\label{sec:triplets}

Let $\mathcal{X} = \left\{x_1, x_2,...,x_n\right\}$ be a set of $n$ objects that we wish to find perceptual similarity amongst.  
We denote the set of possible \emph{triplet queries} 
over $\mathcal{X}$ to be defined as $\mathcal{Q_{\mathcal{X}}} = \left\{\left(a,\left\{b,c\right\}\right)_1,\left(a,\left\{b,c\right\}\right)_2,...\right\}$, where $|\mathcal{Q}_{\mathcal{X}}| = n{n-1\choose 2} = O(n^3)$.
We denote a set of \emph{triplet responses} to such queries (``Object $x_a$ is more similar to $x_b$ than it is to $x_c$.'') as $\mathcal{T} = \left\{\left(a,b,c\right)_1,\left(a,b,c\right)_2,..\right\}$.  
In both cases, we call $x_a$ the ``head'' of the query/response and $x_b$ and $x_c$ the ``pair'' it is being compared to.
For example, the query $\left(1,\left\{2, 3\right\}\right)$ can be answered by either the response $\left(1,2,3\right)$ or $\left(1,3,2\right)$.  
In general, the goal of relative triplet similarity learning methods is to learn a model of similarity over $\mathcal{X}$ that reflects the relationships conveyed through a given $\mathcal{T}$.
%

One such method, Crowd Kernel Learning (CKL), assumes that the underlying perceptual similarity space from which $\mathcal{T}$ is drawn can be modeled by embedding the objects in a ${\hat{d}}$-dimensional space.
CKL learns a representation of $\mathcal{X}$ in the form of an embedding $\hat{\mathbf{X}} = \left\{\hat{\mathbf{x}}_1,\hat{\mathbf{x}}_2,...,\hat{\mathbf{x}}_n\right\} \subset \mathbb{R}^{\hat{d}}$ by solving the following maximization problem:
\begin{equation}
\max_{\hat{\mathbf{X}}}\ \ \hspace{-1em}\sum_{\left(a,b,c\right)\in\mathcal{T}}\hspace{-1em}\log\left(p\left(\left(a,b,c\right)|\hat{\mathbf{X}}\right)\right),
\label{eq:CKL}
\end{equation}
%
%
\begin{equation}
p\left(\left(a,b,c\right)|\hat{\mathbf{X}}\right) = \frac{\mu + d_2^2\left(\hat{\mathbf{x}}_a,\hat{\mathbf{x}}_c\right)}{2\mu + d_2^2\left(\hat{\mathbf{x}}_a,\hat{\mathbf{x}}_c\right) + d_2^2\left(\hat{\mathbf{x}}_a,\hat{\mathbf{x}}_b\right)}.
\label{eq:CKLProb}
\end{equation}
\noindent Here, $d_2^2\left(\cdot,\cdot\right)$ is squared Euclidean distance and $\mu$ is a ``uniqueness'' parameter used to increase stability in the optimization. 
By solving \eqref{eq:CKL}, CKL attempts to position embedded objects $\mathbf{x}_a$ and $\mathbf{x}_b$ closer together than $\mathbf{x}_a$ and $\mathbf{x}_c$ for all triplet responses $\left(a,b,c\right)$.
The probability $p$ can be interpreted as the \emph{likelihood} of a triplet response given an embedding.
As a result, \eqref{eq:CKL} can be viewed as maximizing the log-likelihood of triplet responses in $\mathcal{T}$.
CKL is \emph{nonparametric} in that each object is represented by a unique point in the embedding space, which gives CKL a great deal of freedom in maximizing the likelihood of a given set of triplet responses. 

\subsection{Utilizing Auxiliary Features}
\label{sec:auxFeatures}
A limitation of CKL is that it may not generalize well to certain types of triplet responses yet to be provided.
One way to address this is to include side information about the objects.
In this work, we consider side information in the form of features that characterize each object in some meaningful manner.
Let $\mathbf{X} = \left\{\mathbf{x}_1,\mathbf{x}_2,...,\mathbf{x}_n\right\} \subset \mathbb{R}^d$ be \emph{auxiliary feature} vectors for each object in $\mathcal{X}$.
These features may represent important characteristics in the underlying perceptual similarity space of objects.
If they do, then incorporating them into a learned embedding of the objects may allow it to not only model responses in $\mathcal{T}$, but also generalize to triplet responses yet to be obtained, creating a more complete model of similarity. 
To include this side information, we propose to learn a model comprised of a \emph{parametric} component to be learned over the auxiliary features in conjunction with a nonparametric embedding.
Let $\mathbf{Y} = \left\{\mathbf{y}_1,\mathbf{y}_2,...\mathbf{y}_n\right\} \subset \mathbb{R}^{\hat{d}+d}$ be our combined model of objects, where each object $\mathbf{y}_i$ is represented by a concatenation of vectors:
\begin{equation*}
\mathbf{y}_i = \left[\mb{w}^1\mb{x}_i^1,\mb{w}^2\mb{x}_i^2,..,\mb{w}^d\mb{x}_i^d,\hat{\mb{x}}_i^1,\hat{\mb{x}}_i^2,...,\hat{\mb{x}}_i^{\hat{d}}\right].
\label{eq:combModel}
\end{equation*}
\noindent 
The first $d$ elements of $\mathbf{y}_i$ are the auxiliary features $\mathbf{x}_i$, each weighed by an element of the learned parameter $\mathbf{w}$, i.e. $\mb{w}_i \circ \mb{x}_i$.
A proper setting of $\mathbf{w}$ would emphasize auxiliary features that adhere to the general trend in obtained triplet responses, while reducing the influence of features that do not.
For ease of notation, we denote this parametric component of $\mathbf{y}_i$ as $\mathbf{x}_i'$.  
The last $\hat{d}$ elements of $\mathbf{y}_i$, $\hat{\mathbf{x}}_i$, are free parameters to be learned in a nonparametric fashion similar to traditional CKL.
 
To learn $\mathbf{Y}$ we propose a probabilistic formulation guided by the following intuition.
We assume there are a certain number of characteristics in the true perceptual similarity space from which triplet responses $\mathcal{T}$ are drawn.
A subset of these characteristics are exclusively modeled by $\mathbf{X}$, with $\mathbf{w}$ weighting their importance.
$\hat{\mathbf{X}}$ subsequently models the remaining perceptual characteristics not captured by $\mathbf{X}$ and $\mathbf{w}$.
Hence $\mathbf{w}$ only depends on $\mathbf{X}$ and $\mathcal{T}$, while $\hat{\mathbf{X}}$ depends on $\mb{w}$, $\mb{X}$, and $\mathcal{T}$,
as it must know what was not modeled by the parametric component.
Intuitively, the parametric component leverages auxiliary features to model as much of the perceptual space as possible, while the
nonparametric component effectively fills in the remaining gaps.
 
To learn $\mathbf{w}$ and $\hat{\mathbf{X}}$, we maximize their log-joint posterior given the input variables.  
By our dependence assumptions, we can decompose the log-joint posterior into two terms:
%
%
%
\begin{equation*}
\begin{array}{rl}
\log\left(p\left(\mathbf{w},\hat{\mathbf{X}}|\mathbf{X},\mathcal{T}\right)\right) = & \log\left(p\left(\mathbf{w}|\mathbf{X},\mathcal{T}\right)\right) + \\
& \log\left(p\left(\hat{\mathbf{X}}|\mathbf{w},\mathbf{X},\mathcal{T}\right)\right).
\label{eq:posterior}
\end{array}
\end{equation*}
%
Following the intuition behind our assumptions, we choose to learn $\mathbf{w}$ and $\hat{\mathbf{X}}$ in two stages. 
First, we find $\mathbf{w}$ by maximizing the log-posterior of $\mathbf{w}$ given the input variables.
Then, we fix $\mathbf{w}$ to be the solution in the first step and maximize the log-posterior of $\hat{\mathbf{X}}$ given all other variables.
To learn $\mathbf{w}$, note that the log-posterior of $\mathbf{w}$ can be further decomposed:
\begin{multline*}
\log\left(p\left(\mathbf{w}|\mathbf{X},\mathcal{T}\right)\right) \propto \log\left(p\left(\mathbf{X},\mathcal{T}|\mathbf{w}\right)\right) + \log\left(p\left(\mathbf{w}\right)\right)\\ 
 = \displaystyle \sum_{\left(a,b,c\right)\in\mathcal{T}} \log\left(p\left(\mathbf{X},\left(a,b,c\right)|\mathbf{w}\right)\right) + \log\left(p\left(\mathbf{w}\right)\right).
\label{eq:post2likelihood}
\end{multline*}
%
\noindent By maximizing the log-posterior of $\mathbf{w}$, we maximize the sum of the log-likelihood of $\mathbf{X}$ and $\mathcal{T}$ and a log-prior probability of $\mathbf{w}$.
The log-likelihood of the inputs can be further decomposed into the sum of log-likelihoods of single triplets:
\begin{equation*}
p\left(\mathbf{X},\left(a,b,c\right)|\mathbf{w}\right) = \frac{\mu + d_2^2\left(\mathbf{x}_a',\mathbf{x}_c'\right)}{2\mu + d_2^2\left(\mathbf{x}_a',\mathbf{x}_c'\right) + d_2^2\left(\mathbf{x}_a',\mathbf{x}_b'\right)}.
\label{eq:likelihoodW}
\end{equation*}
These likelihoods are similar to \eqref{eq:CKLProb} with objects represented as a weighing of their auxiliary features. 
We choose the prior on $\mathbf{w}$ to be uniform over $\mathbb{R}^d_+$, effectively restricting $\mathbf{w}$ to be positive, making our maximization problem:
\begin{equation}
\begin{array}{rl}
\displaystyle \max_{\mathbf{w}} & \displaystyle  \hspace{-1em} \sum_{\left(a,b,c\right)\in\mathcal{T}} \hspace{-0.5em} \log\left(p\left(\mathbf{X},\left(a,b,c\right)|\mathbf{w}\right)\right) \\
\mathrm{s.t.} & \mathbf{w} \geq 0.
\end{array}
\label{eq:maxLikelihoodw}
\end{equation}
which we solve via projected gradient descent.

As with $\mathbf{w}$, the log-posterior of $\hat{\mathbf{X}}$ can be decomposed into the sum of the log-likelihood and log-prior.  In this case, the likelihoods over individual triplets are defined as:
\begin{equation*}
p\left(\mathbf{X},\mathbf{w},\left(a,b,c\right)|\hat{\mathbf{X}}\right) = \frac{\mu + d_2^2\left(\mathbf{y}_a,\mathbf{y}_c\right)}{2\mu + d_2^2\left(\mathbf{y}_a,\mathbf{y}_c\right) + d_2^2\left(\mathbf{y}_a,\mathbf{y}_b\right)}.
\label{eq:likelihoodXhat}
\end{equation*}
\noindent 
Here, we choose the prior on $\hat{\mathbf{X}}$ to be uniform over $\mathbb{R}^{\hat{d}}$.  
Thus, to learn the nonparametric portion of our model, we solve the following maximization problem via gradient descent:
\begin{equation}
\max_{\hat{\mathbf{X}}} \sum_{\left(a,b,c\right) \in \mathcal{T}} \log\left(p\left(\mathbf{X},\mathbf{w},\left(a,b,c\right)|\hat{\mathbf{X}}\right)\right).
\label{eq:maxLikelihoodXhat}
\end{equation}
\noindent In solving \eqref{eq:maxLikelihoodw} we learn the parametric portion of the model to fit to the given triplet responses. 
Then, by solving \eqref{eq:maxLikelihoodXhat} with fixed $\mathbf{w}$, we learn the nonparametric component that complements the parametric component.
Because we use a \textbf{T}wo step learning procedure that utilizes \textbf{A}uxiliary features, based  on \textbf{CKL}, we call our method TACKL.

In our formulation, we assume $d$ and $\hat{d}$ are sufficiently small to conform to the common assumption that perceptual similarity is low-dimensional.
For high-dimensional auxiliary features we can apply standard dimensionality reduction techniques, such as PCA, and process this
low-dimensional representation.
Another option is to process the high-dimensional data directly, employing regularization techniques which promote feature sparsity such
as $l_1$-norm regularization~\cite{tibshirani1996regression}.
Such regularization terms may take the place of our prior probabilities; however, since different types of 
priors are not the emphasis of our approach,
we leave these directions as future work.

Formulating similarity learning from triplet responses in terms of probability distributions allows us to reason about triplet queries in an information theoretic context.  
Specifically, we can define concepts such as entropy and information gain.  
In the next section, we show how we use these concepts to actively select the most informative queries to ask humans to learn a more complete model with fewer responses.
%
%

\subsection{Active Learning with Auxiliary Features}
\label{sec:active}
The general strategy we employ for our active learning scheme is to reduce the uncertainty our model has in the position of each object individually.  
As a result, our method works in rounds where $n$ queries are chosen, each with a different head object $x_a$, and actively selected pairs of objects $x_b$ and $x_c$.
Then, for a round $t$, the entire round of queries is posed to a source of feedback, whose responses are then used to learn $\mathbf{w}^{t+1}$ and $\hat{\mathbf{X}}^{t+1}$.
We measure uncertainty in a head $x_a$ at round $t$ by the \emph{entropy} of its learned variables:
\begin{multline*}
H^t\left(\mathbf{w},\hat{\mathbf{x}}_a\right) = \displaystyle \hspace{-0.25em} \int_{\mathbf{w}} \hspace{-0.25em}p\left(\mathbf{w}|\mathbf{X},\mathcal{T}^t_a\right)\log\left(p\left(\mathbf{w}|\mathbf{X},\mathcal{T}^t_a\right)\right) *\\
\displaystyle \hspace{-1em}\int_{\hat{\mathbf{x}}_a} \hspace{-0.4em}p\left(\hat{\mathbf{X}}|\mathbf{w},\mathbf{X},\mathcal{T}^t_a\right)\log\left(p\left(\hat{\mathbf{X}}|\mathbf{w},\mathbf{X},\mathcal{T}^t_a\right)\right)d\hat{\mathbf{x}}_ad\mathbf{w}.
\end{multline*}
\noindent Here, $\mathcal{T}^t_a$ is all obtained triplet responses at round $t$ with head $x_a$. 
In round $t+1$ we wish to select the query for which its response in expectation, when added to $\mathcal{T}^t_a$, will reduce $H^t\left(\mathbf{w},\hat{\mathbf{x}}_a\right)$ the most.  
Thus, we choose the triplet query $\left(a,\left\{b,c\right\}\right)$ with highest expected \emph{information gain}:
\begin{equation*}
H^t\left(\mathbf{w},\hat{\mathbf{x}}_a\right) - p_{abc}^t*H_{abc}^t\left(\mathbf{w},\hat{\mathbf{x}}_a\right) - \left(1-p_{abc}^t\right)*H_{acb}^t\left(\mathbf{w},\hat{\mathbf{x}}_a\right),
\end{equation*}
\noindent where $H_{abc}^t\left(\mathbf{w},\hat{\mathbf{x}}_a\right)$ is $H^t\left(\mathbf{w},\hat{\mathbf{x}}_a\right)$ evaluated over $\mathcal{T}_a^t \cup \left(a,b,c\right) = \mathcal{T}_{abc}^t$, and $p_{abc}^t$ is the probability that a human will respond $\left(a,b,c\right)$ when prompted with $\left(a,\left\{b,c\right\}\right)$:
\begin{equation*}
p_{abc}^t = \int_{\mathbf{w}}p\left(\mathbf{w}|\mathbf{X},\mathcal{T}_t^{abc}\right)\int_{\hat{\mathbf{x}}_a}p\left(\hat{\mathbf{X}}|\mathbf{w},\mathbf{X},\mathcal{T}_t^{abc}\right) d\hat{\mathbf{x}}_ad\mathbf{w}.
\end{equation*}
\noindent For a single head, the first term in the expected information gain is constant.
Thus, our active learning scheme scores pairs of objects given a head using expected query entropy:
\begin{equation}
p_t^{abc}*H^t_{abc}\left(\mathbf{w},\hat{\mathbf{x}}_a\right) + \left(1-p_t^{abc}\right)*H^t_{acb}\left(\mathbf{w},\hat{\mathbf{x}}_a\right),
\label{eq:expEnt}
\end{equation}
\noindent and chooses the one with the lowest score. 
Note, that for finding the entropy, we fix the nonparametric components of all objects besides $x_a$ to be their estimate at round $t$, making entropy a function of only $\hat{\mathbf{x}}_a$ and $\mathbf{w}$. 
A similar assumption is used for A-CKL \cite{tamuz2011adaptively}.
If the prior on their entirely nonparametric model is chosen to be uniform over the current positions of objects, then their scoring criteria amounts to finding information gain if $x_a$ is randomly moved to the exact position of one of the other objects.  
This intuition holds for the nonparametric component of TACKL.
However, the parametric component in our model couples all objects through $\mathbf{w}$.
Thus, when scoring a query, the entropy calculation is based on randomly repositioning $x_a$ via $\mathbf{w}$ and $\hat{\mathbf{x}}_a$, and all other objects through $\mathbf{w}$.
In practice, we perform Monte Carlo integration in \eqref{eq:expEnt} by uniformly sampling $\mathbf{w}$ and $\hat{\mathbf{X}}$ over domains restricted by the magnitudes of their estimates. 

Algorithm \ref{alg:actTACKL} outlines the entire process of actively learning perceptual similarity using our methods.  
The first round (round 0) of query selection is random to establish a basis for subsequent rounds.  
For all rounds $t > 0$, for each head, all pairs of objects are scored.  
For learning similarity over a large number of objects, scoring every triplet is prohibitively expensive.  
In practice, we randomly choose a subset of pairs to score.  
The query that results in the lowest expected entropy for each head is chosen and added to the set of queries for round $t$.  
Then, these queries are posed to the source of feedback for responses, which are added to the running pool. 
\begin{algorithm}[t]
  \center \caption{Active TACKL}
  \label{alg:actTACKL}
  \begin{algorithmic}[1]
    \Require{$\mathbf{X} \subset \mathbb{R}^{d}$}
    \State $\mathcal{T}^0 \gets \emptyset$
    \For{round $t=0,1,2,...$}
    \State $\mathcal{Q}^t \gets \emptyset$
      \For{head $a=1,2,3...,n$}
        \If{$t == 0$}
          \State $b' \gets$ Random draw from $\{1,...n\} - a$
          \State $c' \gets$ Random draw from $\{1,...n\} - \{a,b\}$
        \Else
          \ForAll{$\{b,c\} \subset \{1,...,n\} : b,c \neq a$}
            \State $score\left(a,\left\{b,c\right\}\right) \gets$ \eqref{eq:expEnt} using $\hat{\mathbf{X}}^t$ and $\mathcal{T}^t$          
          \EndFor
          \State $\left\{b',c'\right\} \gets \argmin_{b,c} score\left(a,\left\{b,c\right\}\right)$
        \EndIf
        \State $\mathcal{Q}^t \gets \mathcal{Q}^t \cup \left(a,\left\{b',c'\right\}\right)$ 
      \EndFor
      \State $\mathcal{T}^{t+1} \gets \mathcal{T}^{t} \cup responses\left((\mathcal{Q}^t\right)$
      \State $\mathbf{w}^{t+1} \gets$ solution of \eqref{eq:maxLikelihoodw} using $\mathcal{T}^{t+1}$ and $\mathbf{X}$
      \State $\hat{\mathbf{X}}^{t+1} \gets$ solution of \eqref{eq:maxLikelihoodXhat} using $\mathcal{T}^{t+1}$, $\mathbf{X}$, and $\mathbf{w}^{t+1}$
    \EndFor \\
    \Return $\left\{\mathbf{w}^{t+1},\hat{\mathbf{X}}^{t+1}\right\}$
  \end{algorithmic}
\end{algorithm}
The model parameters are updated from the pool of responses, consequently driving the next round of active selection.

Querying for triplets in rounds has a practical application.
Often to obtain large quantities of human feedback \emph{crowdsourcing} technologies are used.  
For crowdsourcing, human inference tasks are typically deployed in batches so many different people can work on tasks at once.  
By selecting $n$ at a time, one could deploy batches of triplet queries, receive responses, learn a model, and then select $n$ more to repeat the process.
In the next section, we evaluate TACKL and our active variant A-TACKL on both synthetic data as well as real data gained from crowdsourcing triplet responses.
\section{Experiments}
\label{sec:Experiments}
To evaluate TACKL we compare it experimentally to CKL, using both random query selection and their active schemes, A-TACKL and A-CKL.  
The goal is to show the advantage our method gains from using auxiliary information for both random and active query selection schemes.  
We measure the accuracy of a learned perceptual model by \emph{query prediction error}.
For a set of triplet responses, the query prediction error is the number of responses whose likelihood given a model is less than or equal to 50\% divided by the number of responses.
In other words, it is proportional to the number of times the model would predict the incorrect response given a query.
We evaluate each method after each round of query selection to show error as a function of training set size.

In the interest of fairness, we allow the active learning methods to score the same number of randomly selected pairs per head.  
Also, we fix the number of samples in the Monte Carlo integration for A-CKL to be $n\sqrt{n}$, and for $\mathbf{w}$ and $\hat{\mathbf{X}}$ in A-TACKL to be $\sqrt{n}$ and $n$. 
As a result, both active learning methods take the same number of samples.
CKL and TACKL are given the same randomly selected responses each round, and all methods are provided the same random initial round of responses.
We set $\hat{d} \gets d$ for the TACKL methods, $\hat{d}\gets5$ for the CKL methods ($6$ for synthetic experiment), and $\mu \gets 10^{-4}$ for all methods, as slight variations in their settings did not significantly impact any method.
%
\begin{figure}[t]
  \centering 
  \includegraphics[width=\columnwidth]{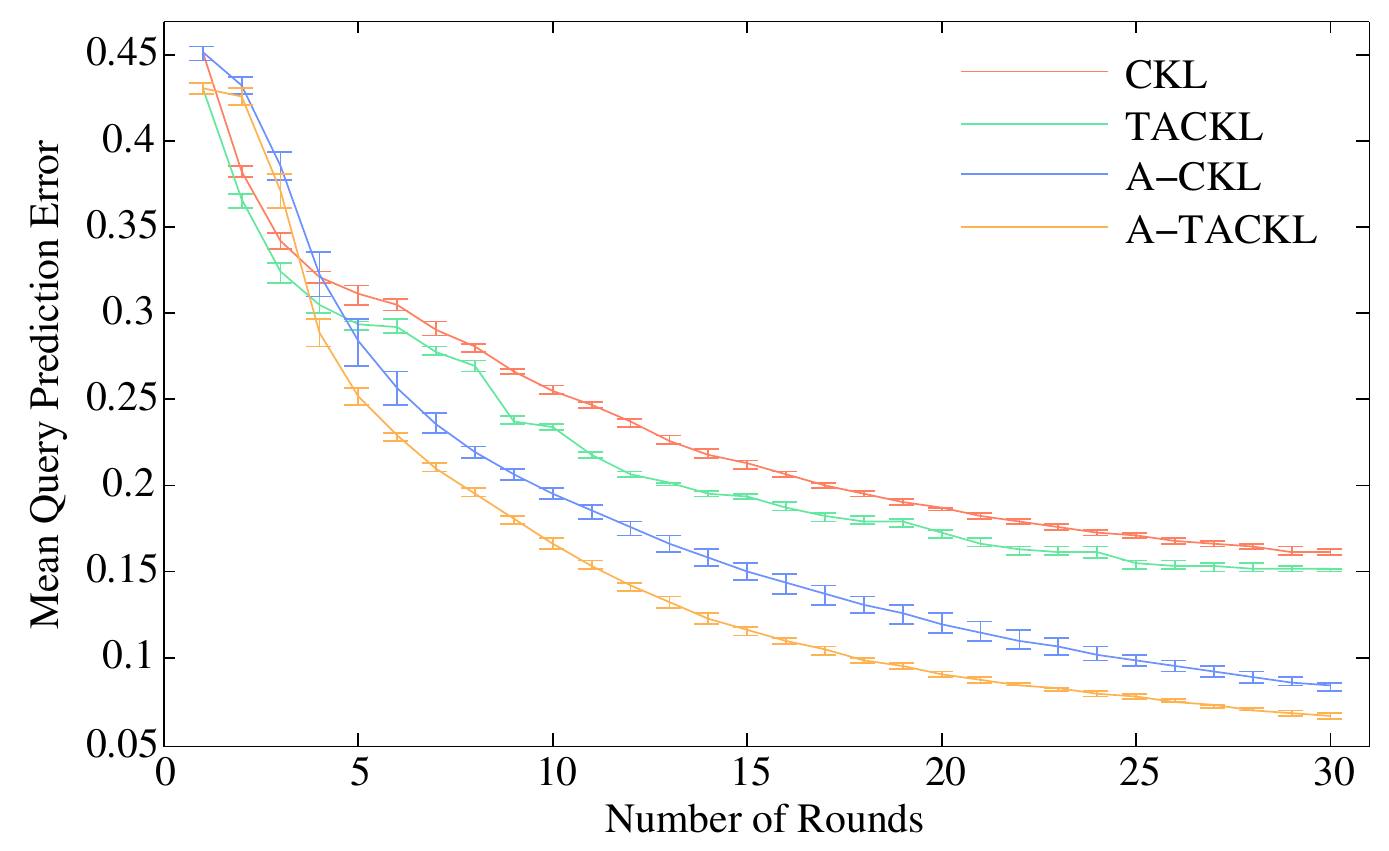}
  \caption{Synthetic experiments (10 trials, 90\% CI)}
  \label{fig:syn}
\end{figure}
\subsection{Synthetic Data}
The first experiment highlights the performance of TACKL in a controlled and idealized synthetic setting.
We generated 250 objects in six dimensions as the ground truth perceptual space $\mathbf{X}^*$, each dimension being drawn from a different distribution:
half from uniform distributions, and half from mixtures of normal distributions.
From $\mathbf{X}^*$ we generate an exhaustive set of all triplet responses.
All queries by the methods are answered by this pool, and the entire pool is used to evaluate query prediction error.
As auxiliary features, the TACKL methods are given three dimensions from the ground truth as well as three more randomly generated noise features.
This input coincides with our model assumptions: potentially only some auxiliary features represent dimensions in the underlying perceptual similarity space.

\textbf{Discussion:} Figure \ref{fig:syn} shows the mean query prediction error versus number of rounds from running experiments for ten trials.  
As with all subsequent figures, error bars represent a 90\% confidence interval. 
Note that both active learning methods perform poorly in the first few rounds.  
With such few triplets in $\mathcal{T}^t_a$, the information gain scores are not meaningful and having variety among pairs seems to allow the random selection methods to learn more accurate models.
However, A-TACKL recovers from this cold start by round four and subsequently has significantly less error than all other methods.
Overall, the TACKL methods are more accurate than their CKL equivalents because they are given auxiliary information that allows for better generalization.  

\subsection{Yummly Food Data}
To test TACKL on real responses obtained from humans, we performed two tests.  
First, we used a dataset of triplet responses gathered over food images sourced from Yummly recipes using AMT~\cite{wilber2014cost}.  
Of the 100 images, 73 have annotations from Yummly measuring to what degree each food image has six taste properties: salty, savory, sour, bitter, sweet, and spicy.
We used these as auxiliary features in our experiment.  
To obtain triplet responses, the authors showed AMT users various-sized grids of food images and were asked to select the images that ``taste most similar''. 
Selected images were deemed the $x_a$ and $x_b$ objects and the rest were deemed $x_c$ for responses $\left(a,b,c\right)$.
Because their goal was to evaluate grid-size versus accuracy of feedback, a certain amount of noise is present in the responses.  
For the 73 images with auxiliary features, the dataset contains 66,497 unique responses, which is roughly 36\% responses to all possible queries. 
To reconcile the lack of a full set of responses to all queries, the active selection methods in our experiment simply chose to ask the highest scoring query for which there was a response.  
\begin{figure}[t]
  \centering 
  \includegraphics[width=\columnwidth]{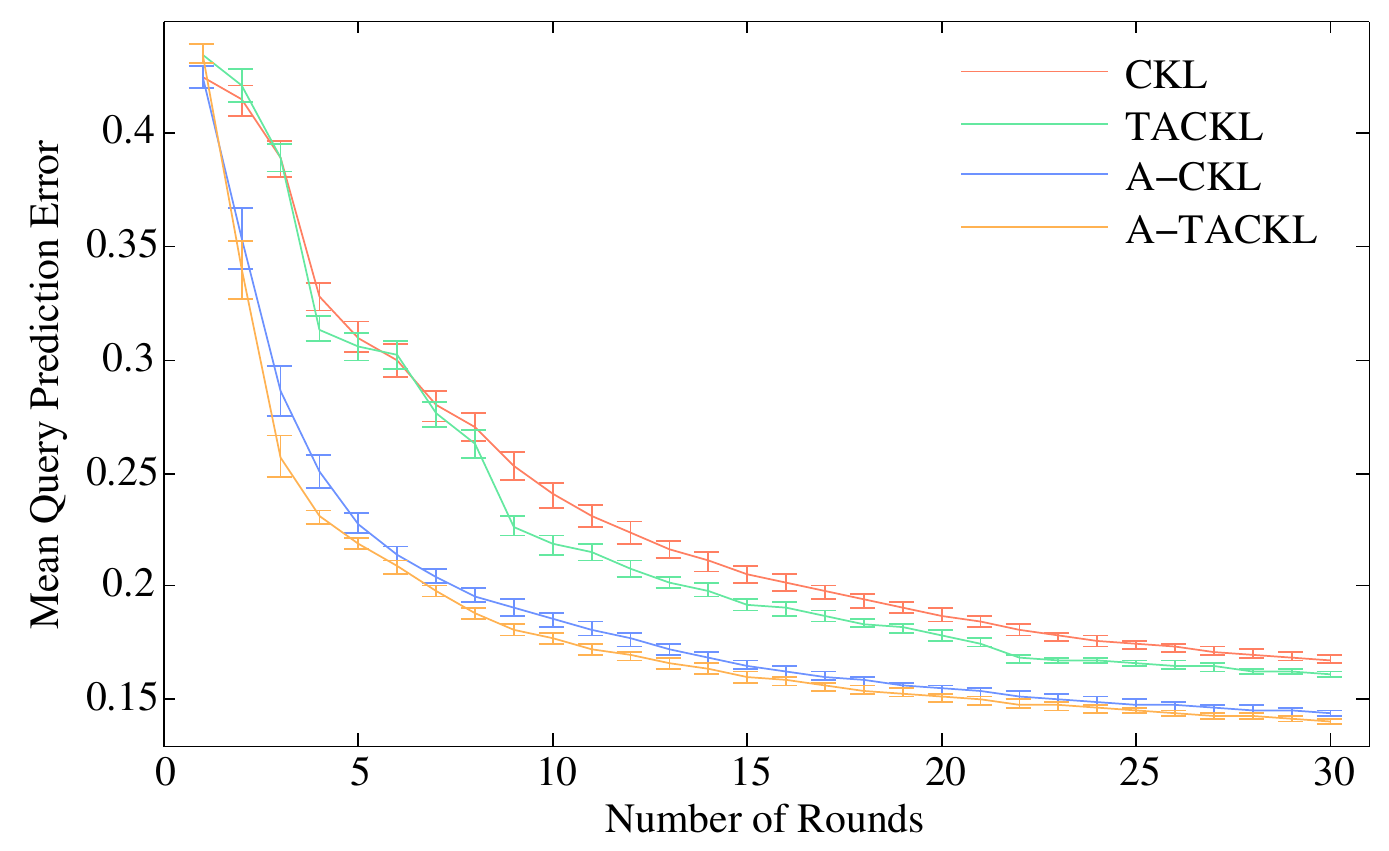}
  \caption{Yummly food data experiments (20 trials, 90\% CI)}
  \label{fig:food}
\end{figure}
We evaluate error for each method on the full set of responses.
\begin{figure*}[t]
  \centering
\begin{subfigure}{0.5\textwidth}
    \centering
    \includegraphics[width=0.95\columnwidth]{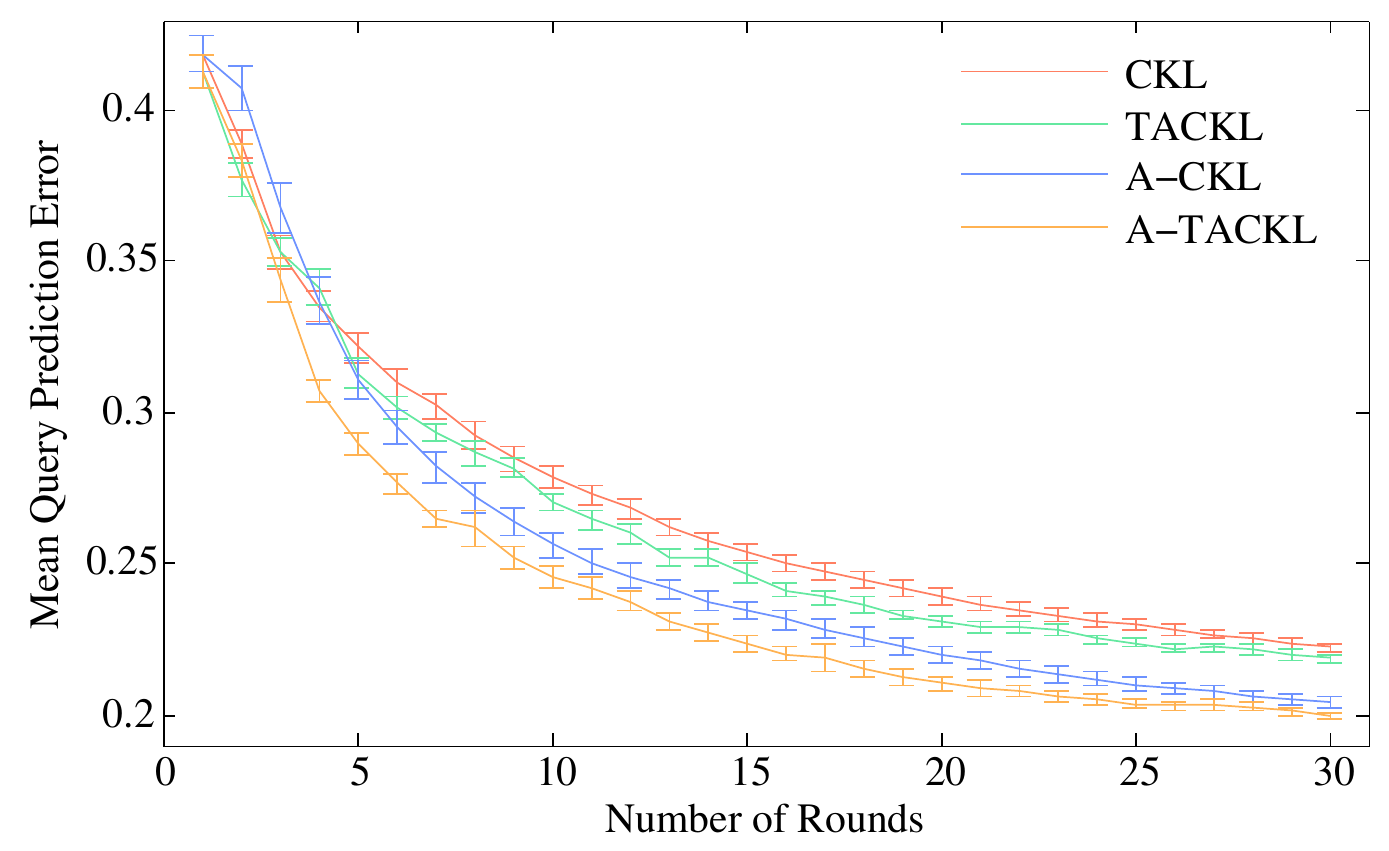}
    \label{fig:Zappos1}
  \end{subfigure}%
\hfill
  \begin{subfigure}{0.5\textwidth}
      \vspace{0.125em}
    \centering
    \includegraphics[width=0.97\columnwidth]{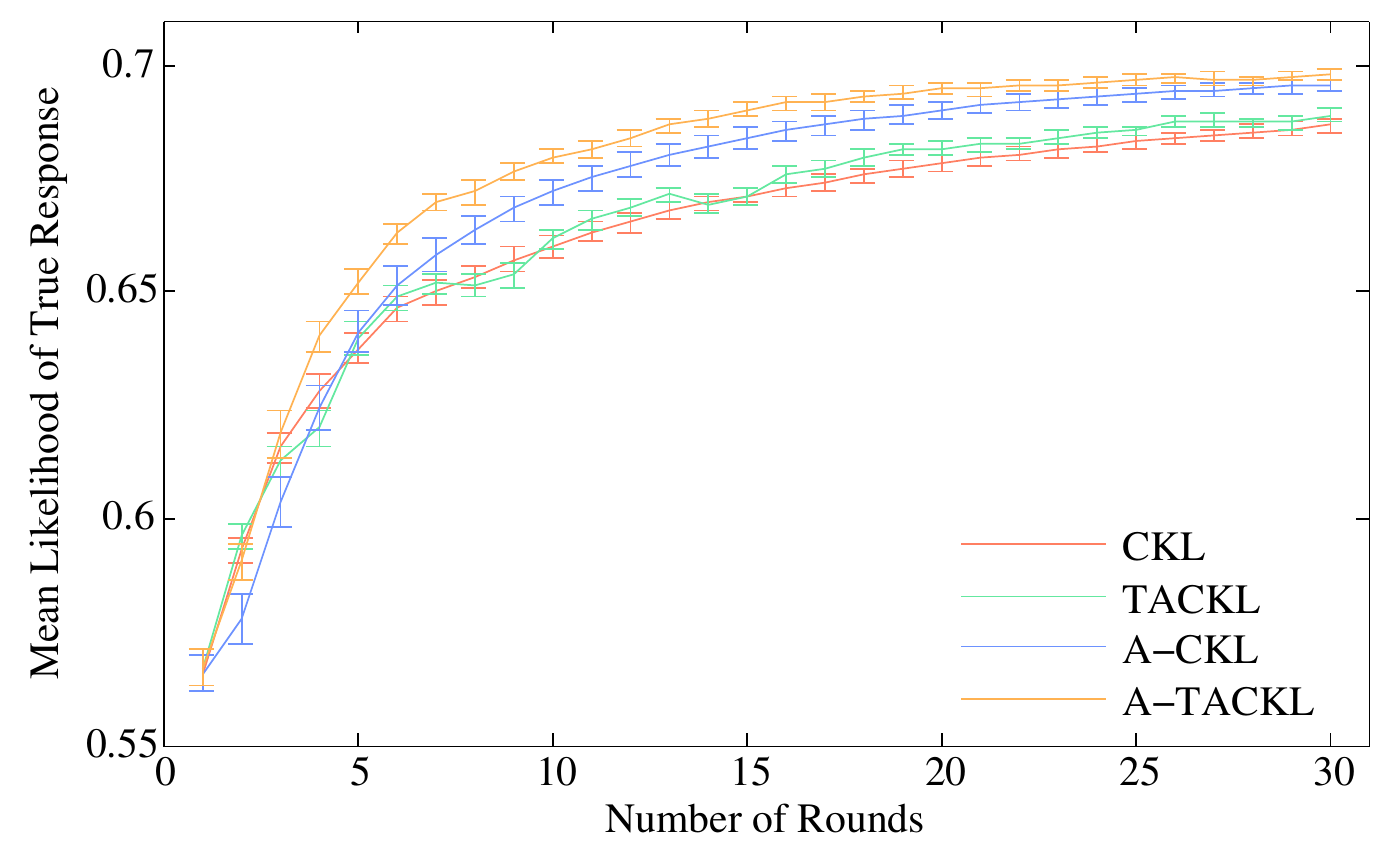}
    \label{fig:Zappos2}
  \end{subfigure}%
    \caption{Zappos shoe data experiments (20 trials, 90\% CI)}
    \label{fig:Zappos}
\end{figure*}

\textbf{Discussion:} The results of our experiments on the Yummly data are shown in Fig. \ref{fig:food}.  
As with the synthetic experiments, the TACKL methods have consistently less error than their CKL counterparts, often by a significant margin.  
In this case, however, the active methods do not suffer from cold start.  
This may be due to the fact that the number of objects and responses to evaluate each model is much less than in the synthetic responses.  
As a result, the active methods can hone in on a satisfactory model to reason about with much fewer responses.  
Also in this experiment, the active methods outperform the random methods by a much larger margin in the middle rounds, indicating that the active methods especially benefited from asking informative queries in early rounds. 
If just the taste features themselves are used as a similarity model, a query prediction error of 0.4562 is achieved.
Despite this, the TACKL methods are able to perform well and find the relevant portions of taste, giving large weight to the ``sweet'' feature and frequently assigning zero weight to ``sour'' and ``spicy''.
Indeed, the data contains many varieties of sweet (pastries, fruits) and not sweet (rice, pasta) foods, while
few of the foods could be considered sour or spicy.

\subsection{Zappos Shoes Data}
To fully evaluate the performance of our active learning approach and to provide a baseline for comparisons with future active triplet learning schemes, we sought to create a dataset containing responses to all possible triplet queries over a nontrivial number of objects. 
By doing so, we could truly evaluate the active triplet query selection methods by allowing them to choose any query and obtain a response from an actual human. 
To our knowledge, there does not currently exist such a dataset that is publicly available. 

To this end, we chose a representative subset of 85 images of shoes from the Zappos50K dataset \cite{yu2014} by performing $k$-means clustering on the images represented by their ImageNet~\cite{krizhevsky2012imagenet} features.  
These images were used in the deployment of an AMT task consisting of all 296,310 triplet queries among images. 
In the supplement, we include a screenshot of the deployed task.    
Each query was dispatched to three different AMT users, and the ground truth response was determined by majority vote. 
Approximately 1,020 different users provided responses to queries.  
Among them, $82\%$ of the queries had full agreement among users, and  $18\%$ had $66\%$ agreement.  
Upon publication of this work, we plan on making this data publicly available.
For auxiliary features, we used the ImageNet feature representations of each image projected to three dimensions using Principal Components Analysis.  
\begin{figure}[t]
  \centering 
  \includegraphics[width=\columnwidth]{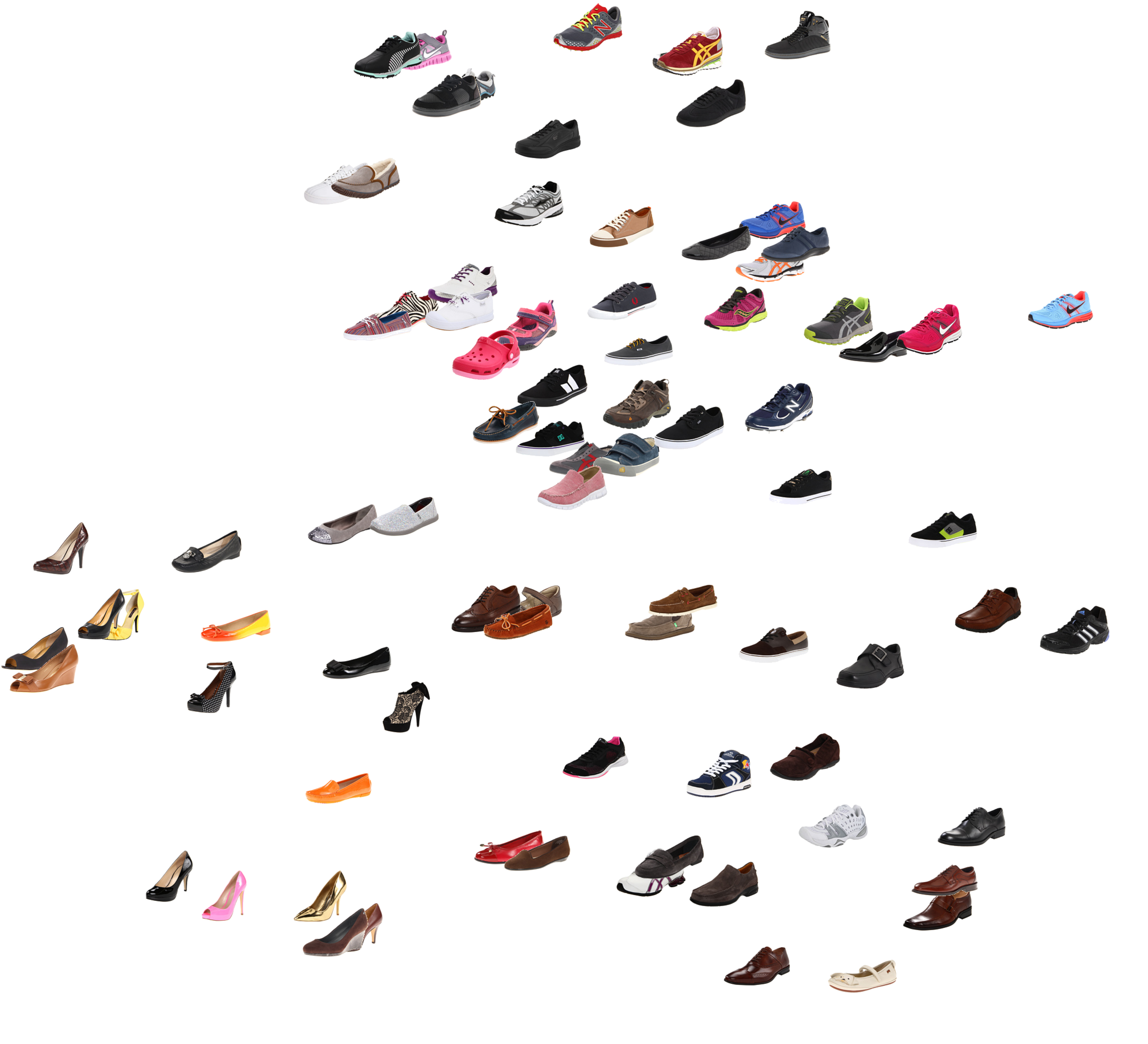}
  \caption{Learned embedding on Zappos Data by A-TACKL}
  \label{fig:embedding}
\end{figure}
\textbf{Discussion:} Figure \ref{fig:Zappos} contains graphs depicting the results of our experiments on the Zappos data.  
The figure on the left shows the mean query error of the learned models as in previous experiments.  
Not shown in the figure is that the full ImageNet features by themselves achieved error of 0.3882.
Most methods surpassed this after two rounds.
The difference between the active and random methods is much smaller.  
This could be attributed to the slight amount of noise in the large pool of responses, which affects both learning model parameters and evaluation of the error.  
As with the Yummly data experiments, A-TACKL does not suffer from cold start.  
However, this is not the case for A-CKL. 
 A-TACKL is able to rely on the auxiliary information when given few triplet responses and learn a model for which active learning benefits.  
Overall, the TACKL methods outperform their CKL counterparts.
The right-most graph in Fig. \ref{fig:Zappos} shows the mean likelihood of all triples responses for all learned models. This metric conveys the \emph{margin} by which the models adhere to all responses.  While TACKL has consistently lower error than CKL, it is much less stable in terms of likelihood.  This implies that TACKL can consistently raise the likelihood of more correct responses above 0.5, but it sometimes does so without raising it far beyond 0.5.  Nevertheless, the TACKL methods perform relatively well according to this metric. By the time A-TACKL is given 30 responses, on average, the distance between $x_a$ and $x_c$ is more than twice than the distance between $x_a$ and $x_b$ for all responses $\left(a,b,c\right)$.  Finally, Fig. \ref{fig:embedding} shows one of the embeddings learned by A-TACKL after 30 rounds, projected to 2 dimensions.


%
                                                                                          
\section{Conclusion and Future Work}
\label{sec:Conclusion}
In this work, we introduced a novel method to learn a model of perceptual similarity from relative triplet responses.  
Our method, TACKL, assumes the likely case that auxiliary features describing objects in question are available.  
With this assumption, TACKL learns an embedding with a parametric component that uses auxiliary information, then a nonparametric component to model true object relationships not captured by the parametric component.  
TACKL is formulated probabilistically, allowing us to create an active learning scheme based on information theoretic notions of which queries are most informative.
In our experimental evaluation, we show that by leveraging auxiliary features and actively selecting triplet queries, TACKL can learn more complete and accurate models of perceptual similarity than methods that do not use side information and select queries at random.

There are two avenues of future work we will pursue.  First, our active learning algorithm works by selecting batches of queries at a time.  
If instead it selected single queries between training phases, further benefits could be gained by updating the model more often between selections.
We will investigate this \emph{online} query selection case further.
Second, a great deal of meta-data is created when obtaining triplet responses via crowdsourcing, such as, who is providing what feedback and which responses have high agreement.
We will investigate the inclusion of this meta-information in designing an active learning framework to gain even further benefits.

\newpage
 
\bibliography{adaptAK}
\bibliographystyle{aaai}
\end{document}